\documentclass{article}

\usepackage{graphicx} 
\usepackage{subfigure}

\usepackage{algorithm}
\usepackage{algorithmic}
\usepackage{appendix}

\usepackage{hyperref}

\newcommand{\BEAS}{\begin{eqnarray*}}
\newcommand{\EEAS}{\end{eqnarray*}}
\newcommand{\BEA}{\begin{eqnarray}}
\newcommand{\EEA}{\end{eqnarray}}
\newcommand{\BEQ}{\begin{equation}}
\newcommand{\EEQ}{\end{equation}}
\newcommand{\BIT}{\begin{itemize}}
\newcommand{\EIT}{\end{itemize}}
\newcommand{\BNUM}{\begin{enumerate}}
\newcommand{\ENUM}{\end{enumerate}}
\newcommand{\BA}{\begin{array}}
\newcommand{\EA}{\end{array}}

\newcommand{\argmin}{\mathop{\rm argmin}}

\newcommand{\tr}{\mathop{ \rm tr}}

\newcommand{\R}{\mathbb{R}}
\newcommand{\BlackBox}{\rule{1.5ex}{1.5ex}}  

\input{macros}

\def\A{{\bf A}}
\def\a{{\bf a}}
\def\B{{\bf B}}

\def\C{{\bf C}}

\def\D{{\bf D}}
\def\d{{\bf d}}

\def\F{{\bf F}}

\def\h{{\bf h}}
\def\K{{\bf K}}
\def\H{{\bf H}}
\def\G{{\bf G}}

\def\R{{\bf R}}
\def\X{{\bf X}}
\def\Y{{\bf Y}}
\def\Q{{\bf Q}}
\def\q{{\bf q}}

\def\S{{\bf S}}
\def\x{{\bf x}}

\def\Z{{\bf Z}}
\def\M{{\bf M}}
\def\m{{\bf m}}

\def\W{{\bf W}}

\def\0{{\bf 0}}
\def\1{{\bf 1}}

\def\RB{{\mathbb R}}

\def\tha{\mbox{\boldmath$\theta$\unboldmath}}

\def\ie{\emph{i.e. }}

\def\argmin{\mathop{\rm argmin}}
\def\tr{\mathsf{tr}}

\def\etal{{\em et al.\/}\,}

\title{A Brief Summary of Dictionary Learning Based Approach for Classification}

\author{
Shu Kong and Donghui Wang\\
\texttt{\{aimerykong, dhwang\}@zju.edu.cn}\\
Institute of Artificial Intelligence,\\
College of Computer Science and Technology, Zhejiang University\\
Hangzhou, China
}

\date{\today}

\begin{document}
\maketitle

\begin{abstract}
\begin{quote}
This note presents some representative methods which are based on dictionary learning (DL) for classification.
We do not review the sophisticated methods or frameworks that involve DL for classification,
such as online DL and spatial pyramid matching (SPM),
but rather, we concentrate on the direct DL-based classification methods.
Here, the ``so-called direct DL-based method'' is the approach directly deals with DL framework by adding some meaningful penalty terms.
By listing some representative methods,
we can roughly divide them into two categories,
\ie (1) directly making the dictionary discriminative and (2) forcing the sparse coefficients discriminative to push the discrimination power of the dictionary.
From this taxonomy,
we can expect some extensions of them as future researches.

\end{quote}
\end{abstract}

\section{Introduction}

Dictionary learning (DL), as a particular sparse signal model,
aims to learn a set of atoms, or called visual words in the computer vision community,
in which a few atoms can be linearly combined to well approximate a given signal.
From the view of compression sensing, it is originally designed to learn an adaptive codebook to faithfully represent the signals with sparsity constraint.
In recent years, researchers have applied DL framework to other applications and achieved state-of-the-art performances,
such as image  denoising~\cite{Denoising_CVPR2006} and inpainting~\cite{Elad_ProcIEEE2010}, clustering~\cite{L1_graph,Wright_ProcIEEE2010}, classification~\cite{Bradley_NIPS2008,SupervisedDL}, etc.

It is well-known that the conventional DL framework is not adapted to classification as a result that the learned dictionary is merely used for signal reconstruction.
Therefore, to circumvent this problem,
researchers have developed several approaches to learn a classification-oriented dictionary in a supervised learning fashion by
exploring the label information.
In this note,
we review the some existing representative DL-based classification methods.
Through comparison,
we can roughly divide them into two categories:
(1) directly forcing the dictionary discriminative,
or (2) making the sparse coefficients discriminative (usually through simultaneously learning a classifier) to promote the discrimination of the dictionary.
The first category, named Track I in this note,
mainly uses representation error for the final classification,
whereas, the second category (Track II) can utilize the sparse coefficients as new feature representation for classification.

Track I includes Meta-face learning~\cite{metaface_ICIP2010} and DL with structured incoherence~\cite{StructuredIncoherenceSharedFeatures_DL},
and Track II contains supervised DL~\cite{SupervisedDL}, discriminative K-SVD~\cite{Discriminative_KSVD}, label consistence K-SVD~\cite{LabelConsistent_KSVD} and Fisher discrimination DL~\cite{FisherDL_ICCV2011}.
The abbreviations of these methods are listed in Table~\ref{tab:abbreviation}.

The organization of this note is as follows.
In the end of this section,
we review an important method called sparse representation-based classification~\cite{SRC},
then introduce the general dictionary learning framework with notations used in this note.
Note that even though SRC do not learn dictionaries, it opens the prologue of classification based on sparse coding technique.
In Section~\ref{sec:Track1},
we introduce Meta-face learning~\cite{metaface_ICIP2010} and DLSI~\cite{StructuredIncoherenceSharedFeatures_DL} as two specific examples of Track I,
which uses the reconstruction error for the final classification like what SRC does.
Its counterpart, \ie Track II, will presented in Section~\ref{sec:Track2},
including SupervisedDL~\cite{SupervisedDL},
D-KSVD~\cite{Discriminative_KSVD},
LC-KSVD~\cite{LabelConsistent_KSVD} and FisherDL~\cite{FisherDL_ICCV2011}.
In Section~\ref{sec:summary},
we give a brief summary on DL-based classification methods,
and expect some extensions in the future work.

\begin{table}
\caption{Two categories of DL-based classification methods.}
\begin{center}
\begin{tabular}{ l | c }
  \hline\hline
  Category & Representative Approaches\\
  \hline
  {Track I}  & Meta-face learning~\cite{metaface_ICIP2010}, DLSI~\cite{StructuredIncoherenceSharedFeatures_DL}\\
  {Track II} & SupervisedDL~\cite{SupervisedDL}, D-KSVD~\cite{Discriminative_KSVD}, LC-KSVD~\cite{LabelConsistent_KSVD}, Fisher DL~\cite{FisherDL_ICCV2011}\\
  \hline
  \hline
\end{tabular}
\end{center}
\label{tab:abbreviation}
\end{table}

\subsection{Sparse Representation-Based Classification}
Wright \etal~\cite{SRC} propose the sparse representation based classification (SRC) method for robust face recognition,
and achieve very impressive results.
Suppose there are $C$ classes of individual faces,
let $\D=[\X_1,\dots,\X_c,\dots,\X_C] \in \RB^{d\times N}$ be the set of original training samples,
where $\X_c \in \RB^{d\times N_c}$ is the sub-set of all the $N_c$ vector-represented training samples from class $c$.
SRC treats the original data set as an overall dictionary.
Denote by $\x \in \RB^{d}$ a query facial image,
then SRC identifies $\x$ as the following two-stage procedure:
\begin{enumerate}
  \item sparsely code $\x$ over $\X$ via $\ell_1$-norm minimization
            \begin{equation}
            \a = \argmin_{\a} \Vert\x-\D\a\Vert_2^2+\lambda\Vert\a\Vert_1,
            \end{equation}
            where $\lambda$ is a scalar constant.
  \item identify $\x$ to the $c^{th}$ class that
            \begin{equation}
            c = \argmin_{i} \Vert\x-\X_i\delta_i(\a)\Vert_2^2,
            \end{equation}
            where $\delta_i(\cdot)$ is a vector indicator function that extract the elements corresponding to the $i^{th}$ class.
\end{enumerate}

SRC achieves very impressive performance in face recognition, and robust to noises such as occlusion, lighting, etc.
Even if SRC learns no dictionaries for classification,
it acts as one vanguard to open the prologue of classification with the help of sparse coding.
In this view, we can see SRC naively uses all the training samples as one dictionary,
wherein the class-specific training sets are sub-dictionaries contributing to discrimination.

\subsection{Dictionary Learning Framework}
Learning an adaptive dictionary (possible overcomplete) aims to provide a basis pool in which a few bases can be linearly combined to approximate a novel signal.
Suppose there are a set of signals, denoted by $\X=[\x_1,\dots,\x_i,\dots,\x_N]$,
where $\x_i$ is the $i^{th}$ signal.
Then the conventional dictionary learning framework learns the dictionary as below:
\begin{equation}
\label{equa:DL2}
\begin{split}
\{ \A, \D\} &= \argmin\limits_{%
    \substack{
    \D\in\RB^{d\times K}\\
    \A \in\RB^{K \times N} }%
}
\sum_{i=1}^{N}
\Vert \x_{i} - \D\a_{i}\Vert_{2}^{2} + \lambda \Vert \a_i \Vert_{1}\\
&=\argmin\limits_{%
    \substack{
    \D\in\RB^{d\times K}\\
    \A \in\RB^{K \times N} }%
}
\Vert \X - \D\A\Vert_{F}^{2} + \lambda \Vert\A\Vert_1 \\
&\text{s.t. $\Vert\d_{i}\Vert_{2}^{2} \le 1$, for $\forall i=1,\dots,N $,}
\end{split}
\end{equation}
where $\A=[\a_{1},\dots,\a_{N}]$ is the coefficient matrix and
$\Vert\A\Vert_1 = \sum_{i}^{N} \Vert\a_i\Vert_1$.

It is widely known that classic dictionary learning framework is designed for a reconstruction task instead of classification tasks, even if good classification results are achieved in the literature. It is believed that classification performance will be further improved if we carefully learn a classification-oriented dictionary.
In next section,
we will have a look at several DL-based classification methods belonging to Track I.

\section{Track I: Directly Making the Dictionary Discriminative}
\label{sec:Track1}

The methods from Track I use the reconstruction error for the final classification,
thus the learned dictionary ought to be as discriminative as possible.
Inspired by SRC,
Yang \etal propose meta-face learning~\cite{metaface_ICIP2010} to learn an adaptive dictionary for each class,
and Ramirez \etal add a sophisticated term to derive more delicate classification-oriented dictionaries.
Now, we present the two methods.

\subsection{Meta-Face Learning}
SRC directly adopts the original facial images as the dictionary,
however, as discussed in~\cite{metaface_ICIP2010},
this pre-defined dictionary will incorporate much redundancy as well as noise and trivial information that can can be negative to the face recognition.
Additionally, when the training data grows, the computation of sparse coding will become a main bottleneck.
Focusing on this problem,
Yang \etal~\cite{metaface_ICIP2010} propose a Metaface learning method to learn a class-specific dictionary for each object:
\begin{equation}
\begin{split}
\D_i &= \argmin_{\D_i} \Vert\X_i-\D_i\A_i\Vert_2^2+\lambda\Vert\A_i\Vert_1,\\
& \text{s.t. } \Vert \d^i_j \Vert_2 \le 1, \forall j = 1,\dots,K,
\end{split}
\label{equa:dl}
\end{equation}
where matrix $\X_i \in \RB^{d\times N_i}$ contains all the training images from the $i^{th}$ class as its columns,
$\d^i_j$ is the $j^{th}$ column of the $i^{th}$ class-specific sub-dictionary $\D_i=[\d^i_1,\dots,\d^i_K]\in \RB^{d\times K}$,
and $\Vert\A_i\Vert_1 $ is defined as the summation of $\ell_1$-norm of all the columns of $\A_i=[\a_1^i,\dots,\a_{N_i}^i]\in \RB^{K \times N_i}$,
\ie $\Vert\A_i\Vert_1 = \sum_{j}^{N_i} \Vert\a^i_j\Vert_1$.
Metaface learning method concatenates all the sub-dictionaries as an overall dictionary $\D = [\D_1,\dots,\D_C]$ for classification,
the same as the second stage of SRC.

\subsection{Dictionary Learning with Structured Incoherence }
Ramirez \etal note that the learned sub-dictionaries may share some common bases,
\ie some visual words from different sub-dictionaries can be very coherent
\cite{StructuredIncoherenceSharedFeatures_DL}.
Undoubtedly, the coherence of the atoms can be used for reconstructing the query image interchangeably,
and the reconstruction error based classifier will fail in identifying some queries.
To circumvent this problem,
they add an incoherence term term to drive the dictionaries associated to different classes as independent as possible.

The incoherence term is denoted as ${\mathcal Q}(\D_{i},\D_{j}) = \Vert \D^{T}_i \D_{j}\Vert_{F}^{2}$.
It is easy to see this term drives the atoms from different sub-dictionaries to be as independent/incoherent as possible.
Therefore, Ramirez \etal derive the final dictionary learning method with structured incoherence as below:
\begin{equation}
\label{equa:IncoherentDL}
\begin{split}
\min\limits_{\{\D_i,\A_i\}_{i=1,\dots,C}}
&\sum\limits_{i=1}^{C} \Biggl\{ \Vert \X_{i} - \D_{i}\A_{i} \Vert_{F}^{2} + \lambda \Vert \A_{i} \Vert_{1} \Biggr\}
+ \eta \sum\limits_{i \not= j}\Vert \D_{i}^{T}\D_{j}\Vert_{F}^{2},
\end{split}
\end{equation}
where $\A_{i} = [\a_{i}^{1},\dots,\a_{i}^{n_i}] \in \RB^{k_i \times n_i}$, each column $\a_{i}^{j}$ is the sparse code corresponding to the signal $j \in [1,\dots,n_i]$ in class $i$.

They empirically note that even though the incoherence term is imposed in the dictionaries,
atoms representing common features in all classes tend to appear repeated almost exactly in dictionaries corresponding to different classes~\cite{StructuredIncoherenceSharedFeatures_DL}.
Being so common, these atoms are used often and their associated reconstruction coefficients have a high absolute value $|\a_{r}|$,
$r \in \{1,\dots,k_i\}$,
thus making the reconstruction costs similar.
They further propose to detect such atoms is to inspect the already available $\D_{i}^{T}\D_{j}$ matrices,
whose absolute values represent the inner products between atoms.
By ignoring the coefficients associated to these common atoms when computing the reconstruction error,
they improve the discriminatory power of the system.

\section{Track II: Making the Coefficients Discriminative}
\label{sec:Track2}

Track II is different from Track I in the way of discrimination.
Contrary to Track I, it forces the sparse coefficients to be discriminative,
and indirectly propagates the discrimination power to the overall dictionary.
Track II only need to learn an overall dictionary,
instead of class-specific dictionaries.
In this section,
we list several recent-proposed methods belonging to Track II.

\subsection{Supervised Dictionary Learning}
Before presenting this method,
we have to clarify that the Supervised DL (SupervisedDL) method is a specific approach proposed in \cite{SupervisedDL},
regardless of other possible supervised DL framework.

Mairal \etal propose to combine the logistic regression with conventional dictionary learning framework as below:
\begin{equation}
\label{equa:SupervisedDL}
\begin{split}
( \A, \D) &= \argmin\limits_{%
    \substack{
    \tha \\
    \D\in\RB^{d\times K}\\
    \A\in\RB^{K \times N} }%
}
\sum\limits_{i=1}^{N}
({\mathcal C}(y_{i} f(\x_{i},\a_{i},\tha)) + \lambda_{0}\Vert \x_{i} - \D\a_{i}\Vert_{2}^{2} + \lambda_1 \Vert \a_i \Vert_{1})
+ \lambda_{2} \Vert \tha \Vert_{2}^{2},\\
&\text{s.t. $\Vert\d_{i}\Vert_{2}^{2} \le 1$, for $\forall i=1,\dots,N $,}
\end{split}
\end{equation}
where ${\mathcal C}$ is the logistic loss function (${\mathcal C}(x)=\log(1+e^{-x})$),
which enjoys properties similar to that of the hinge loss from the SVM literature,
while being differentiable,
and $\lambda_2$ is a regularization parameter which prevents overfitting.
This is the approach chosen in~\cite{SelfTaught}.
And $f$ is a classification function --- linear in $\a$: $f(\x, \a, \tha) = \tha^{T}\a+b$ wherein $\tha \in \RB^{K}$,
or bilinear in $\a$ and $\x$: $f(\x,\a,\tha) = \x^{T}\W\a+b$ wherein $\tha = \{ \W \in \RB^{d\times K}, b \in \RB \}$.



\subsection{Discriminative K-SVD for Dictionary Learning}
Zhang and Li propose discriminative K-SVD (D-KSVD) to simultaneously achieve a desired dictionary which has good representation power while supporting optimal discrimination of the classes~\cite{Discriminative_KSVD}.
D-KSVD adds a simple linear regression as a penalty term to the conventional DL framework:
\begin{equation}
\label{equa:D_KSVD}
\begin{split}
(\D,\W,\A) & = \argmin\limits_{\D,\W,\A} \Vert \X - \D\A \Vert_{F}^{2} + \lambda_{1}\Vert \H-\W\A \Vert_{F}^{2}
+ \lambda_{2} \Vert\A \Vert_{1} + \lambda_{3} \Vert \W \Vert_{F}^{2},
\end{split}
\end{equation}
where $\H = [\h_1,\dots,\h_N] \in \RB^{C\times N}$ is the label of the training images, in which $\h_{n}=[0,\dots,0,1,0,\dots,0]$: the position of non-zero element indicates the class. And $\W$ is the parameter of the classifier, $\lambda_{1}$, $\lambda_{2}$ and $\lambda_{3}$ are scalars controlling the relative contribution of the corresponding terms.

Note that the first two terms can be fused into one,
and the term $\Vert \W \Vert_{F}^{2}$ can be dropped during computation owing to the protocol of the original K-SVD algorithm(details in~\cite{Discriminative_KSVD}).
After obtaining the classifier parameter $\W$ and the dictionary,
the final classification can be very fast for a query image.

\subsection{Label Consistent K-SVD}

Jiang \etal propose a label consistent K-SVD (LC-KSVD) method to learn a discriminative dictionary for sparse coding~\cite{LabelConsistent_KSVD}.
They introduce a label consistent constraint called ``discriminative sparse-code error", and combine it with the reconstruction error and the classification error to form a unified objective function as below:
\begin{equation}
\label{equa:LC2_KSVD}
\begin{split}
(\D,\W,\A) & = \argmin\limits_{\D,\W,\A} \Vert \X - \D\A \Vert_{F}^{2} + \lambda_{1}\Vert \Q-\G\A \Vert_{F}^{2}
+ \lambda_{2}\Vert \H-\W\A \Vert_{F}^{2} + \lambda_3 \Vert \A\Vert_{1}\\
&\text{s.t. $\Vert\d_{i}\Vert_{2}^{2} \le 1$, for $\forall i=1,\dots,N $,}
\end{split}
\end{equation}
where $\H$ and $\W$ are the same as that of D-KSVD described in the previous subsection,
$\Q = [\q_{1},\dots,\q_{N}] \in \RB^{K\times N}$ is the label consistence term.
Here $\q_{n}=[0,\dots,1,\dots,1,0,\dots,0]^T \in \RB^{K}$ is an indicator corresponding to the input signal $\x_{n}$ from suitable class:
the non-zero values of $\q_{n}$ occur at those indices where the input signal $\x_{n}$ and the dictionary codeword $\d_{k}$ share the same label.

The term $\Vert \Q-\G\A \Vert_{F}^{2}$ represents the discriminative sparse-code error, which enforces that the sparse codes $\A$ approximate the discriminative sparse codes $\Q$. It forces the signals from the same class to have very similar sparse representations, \ie encouraging label consistency in resulting sparse codes.
At the same time,
the linear regression term $\Vert \H-\W\A \Vert_{F}^{2}$ is added, which is the same as that of D-KSVD~\cite{Discriminative_KSVD}.
Intuitively, the final classification mechanism is very fast owing to the classifier parameter matrix $\W$.

\subsection{Fisher Discriminant Dictionary Learning}
Yang \etal propose Fisher discrimination dictionary learning (FisherDL) method based on the Fisher criterion to learn a structured dictionary~\cite{FisherDL_ICCV2011},
whose atom has correspondence to the class label.
The structured dictionary is denoted as $\D = [\D_1,\dots,\D_C]$,
where $\D_c$ is the class-specific sub-dictionary associated with the $c^{th}$ class.
Denote the data set $\X = [\X_1,\dots,\X_{C}]$,
where $\X_{c}$ is the sub-set of the training samples from the $c^{th}$ class.
Then they solve the following formulation over the dictionary and the coefficients to derive the desired discriminative dictionary:
\begin{equation}
\label{equa:FDDL}
\begin{split}
(\D, \A)&= \argmin\limits_{%
    \substack{
    \D\in\RB^{d\times K}\\
    \A\in\RB^{K \times N} }%
}
{\mathcal C}(\X, \D, \A) + \lambda_{1} \Vert \A \Vert_{1} + \lambda_{2} f(\A),\\
&\text{s.t. $\Vert\d_{i}\Vert_{2}^{2} \le 1$, for $\forall i=1,\dots,N $,}
\end{split}
\end{equation}
where ${\mathcal C}(\X,\D,\A)$ is the discriminative fidelity term (pending to discuss it as below);
$\Vert \A \Vert_1$ is the sparsity constraint;
$f(\A)$ is a discrimination constraint (as discussed below) imposed on the coefficient matrix $\A$.

\textbf{The discriminative fidelity term} We can write $\A_{i}$, the representation of $\X_{i}$ over $\D$, as $\A_{i} = [\A_{i}^{1};\dots;\A_{i}^{c};\dots;\A_{i}^{C}]$, where $\A_{i}^{c}$ is the coding coefficient of $\X_{i}$ over the sub-dictionary $\D_{c}$.
Denote the representation of $\D_{c}$ to $\X_{i}$ as $\R_{c} = \D_{c}\A_{i}^{c}$.
First of all, the dictionary $\D$ should be able to well represent $\X_{i}$, and there is $\X_{i} \approx \D\A_{i} = \D_{1}\A_{i}^{1}+\dots+ \D_{j}\A_{i}^{j}+\dots+ \D_{C}\A_{i}^{C}=\R_{1}+\dots + \R_{j} + \dots + \R_{C}$.
Second, since $\D_i$ is associated with the $i^{th}$ class, it is expected that $\X_{i}$ should be well represented by $\D_i$ but not by $\D_j$, $j\not= i$. This implies that $\A^{i}$ should have some significant coefficients such that $\X_{i}-\D_{i}\A_{i}^{i}$ is small, while $\A_{i}^{j}$ should have nearly zero coefficients such that $\D_{j}\A_{i}^{j}$ is small. Thus the discriminative fidelity term is defined as:
\begin{equation}
\label{equa:DFterm_FDDL}
\begin{split}
{\mathcal C}(\X_i, \D, \A_i) = & \Vert \X_{i} - \D \A_{i}\Vert_{F}^{2} + \Vert \X_{i} - \D_{i} \A_{i}^{i}\Vert_{F}^{2}
+ \sum\limits_{j\not=i}\Vert \D_{j}\A_{i}^{j} \Vert_{F}^{2},\\
\end{split}
\end{equation}

\textbf{The discriminative coefficient term} To make dictionary $\D$ be discriminative for the samples in $\X$, we can make the coding coefficient of $\X$ over $\D$, \ie $\A$, be discriminative. Based on Fisher Criterion, this can be achieved by minimizing the within-class scatter of $\A$, denoted by $\S_{W}$ and maximizing the between-class scatter of $\A$, denoted by $\S_{B}$. $\S_{W}$ and $\S_{B}$ are defined as:
\begin{equation}
\begin{split}
\S_{W}& = \sum\limits_{c=1}^{C}\sum\limits_{\x_{i}\in\X_{c}}(\a_{i} - \m_{c})(\a_{i} - \m_{c})^{T}\\
\S_{B}& = \sum\limits_{c=1}^{C}(\m_{c} - \m)(\m_{c} - \m)^{T}
\end{split}
\nonumber
\end{equation}

Intuitively, we can define $f(\A)$ as $\tr(\S_{W})-\tr(\S_B)$. However, such an $f(\A)$ is non-convex and unstable. To solve this problem, we propose to add an elastic term $\Vert \A\Vert_{F}^{2}$ into $f(\A)$:
\begin{equation}
\label{equa:DCterm_FDDL}
\begin{split}
f(\A) &= \tr(S_{W}) - \tr(S_B) + \eta \Vert \A \Vert_{F}^{2}
\end{split}
\end{equation}

Incorporating all the terms, we have the following FDDL model:
\begin{equation}
\label{equa:FDDL_detail}
\begin{split}
(\D, \A)&= \argmin\limits_{\D,\A}
\left\{%
\begin{array}{ll}
\sum\limits_{c=1}^{C}{\mathcal C}(\X_{i}, \D, \A_{i})+
\lambda_2(\tr(S_{W}) - \tr(S_B) + \eta \Vert \A \Vert_{F}^{2})
+ \lambda_{1}\Vert \A\Vert_{1}
\end{array}
\right\}
\end{split}
\end{equation}

There are some crucial issues related to their model,
such as the convexity of $f(\A)$ and sparse coding,
and they discuss these issue in depth~\cite{FisherDL_ICCV2011}.
As for classification,
they still utilize the reconstruction error as that of Track I.

\section{Summary}
\label{sec:summary}

In previous two sections, we review some representative DL-based classification approaches, both from Track I and Track II.
Obviously, it is intuitive but effective to add some sophisticated discrimination term to the conventional DL framework to derive a well-learned dictionary for classification.

If we check these methods,
we can anticipate a general framework here:
\begin{equation}
\label{equa:general}
\begin{split}
\min\limits_{\D, \W, \A}& {\mathcal C}(\Y, \X, \D, \A)
+ \eta f(\W,\A,\Y)
+ \lambda_{\A} h_{\A}(\A)
+ \lambda_{\W} h_{\W}(\W) \\
&\text{s.t. constraint on $\D$,}
\end{split}
\end{equation}
where ${\mathcal C}(\Y, \X, \D, \A) $ is the conventional DL framework,
$f(\W,\A,\Y)$ is the discrimination term on the sparse coefficients,
$h_{\A}$ and $h_{\W}$ are the Lagrange constraints on the sparse coefficient matrix  $\A$ and the projector $\W$,
$\eta$ and $\lambda$'s are scalars to balance their weights.
Note $\W$ does not necessarily mean only one projector, but rather represents several ones.
From Eq.~\ref{equa:general},
we can see that, by employing the label matrix $\Y$,
the discriminative dictionary can be learned directly in the term ${\mathcal C}(\Y, \X, \D, \A)$,
at the same time,
the term $f(\W,\A,\Y)$ can also propagate the discrimination power of the coefficients to the dictionary,
making the dictionary even more discriminative and reliable for classification.
Obviously, if we set $\eta=0$, Eq.~\ref{equa:general} degrades to Track I;
if we omit the label information in term ${\mathcal C}(\Y, \X, \D, \A)$,
Eq.~\ref{equa:general} degenerates to Track II.
Note that FisherDL~\cite{FisherDL_ICCV2011} can also be cast as a specific example of Eq.~\ref{equa:general},
which drives the dictionary to be as discriminative as possible from two directions (direct push and indirect push by the coefficients).

Besides,
the main concern seems to be the trade-off between the classification accuracy and the complexity of formulation.
Furthermore,
when meeting large scale database,
these methods will be time consuming in learning the dictionary.
Therefore, how to extend these method to online version is an interesting but significant research.

\bibliography{bib}
\bibliographystyle{ieee}

\end{document}